# Assessing airborne laser scanning and aerial photogrammetry for deep learning-based stand delineation

Håkon Næss Sandum[1*], Hans Ole Ørka[1], Oliver Tomic[2], and Terje Gobakken[1]

[1] Faculty of Environmental Sciences and Natural Resource Management, Norwegian University of Life Sciences, NMBU, P.O. Box 5003, NO-1432 Ås, Norway.

[2] Faculty of Science and Technology, Norwegian University of Life Sciences, NMBU, P.O. Box 5003, NO-1432 Ås, Norway.

* hakon.nass.sandum@nmbu.no

Abstract

Accurate forest stand delineation is essential for forest inventory and management but remains a largely manual and subjective process. A recent study has shown that deep learning can produce stand delineations comparable to expert interpreters when combining aerial imagery and airborne laser scanning (ALS) data. However, temporal misalignment between data sources limits operational scalability. Canopy height models (CHMs) derived from digital photogrammetry (DAP) offer better temporal alignment but may smoothen canopy surface and canopy gaps, raising the question of whether they can reliably replace ALS-derived CHMs. Similarly, the inclusion of a digital terrain model (DTM) has been suggested to improve delineation performance, but has remained untested in published literature. Using expert-delineated forest stands as reference data, we assessed a U-Net-based semantic segmentation framework with municipality-level cross-validation across six municipalities in southeastern Norway. We compared multispectral aerial imagery combined with (i) an ALS-derived CHM, (ii) a DAP-derived CHM, and (iii) a DAP-derived CHM in combination with a DTM. Results showed comparable performance across all data combinations, reaching overall accuracy values between 0.90-0.91. Agreement between model predictions was substantially larger than agreement with the reference data, highlighting both model consistency and the inherent subjectivity of stand delineation. The similar performance of DAP-CHMs, despite the reduced structural detail, and the lack of improvements of the DTM indicate that the framework is resilient to variations in input data. These findings indicate that large datasets for deep learning-based stand delineations can be assembled using projects including temporally aligned ALS data and DAP point clouds.

# Introduction

A forest stand can be defined as a relatively homogeneous community of trees that exhibits sufficient uniformity in its attributes to be distinguishable from adjacent communities (Smith 1986). These characteristics encompass species composition, size class distribution, age, stocking level, and the spatial arrangement of trees reflecting historical and local silvicultural practices, as well as site characteristics such as topography and site index (Baker 1934, Husch et al. 1993). Consequently, a stand constitutes a relatively homogenous area suited to a specific management regime and serves as the fundamental unit for inventory, forest management, and financial analysis. Furthermore, stand-level estimates are used as input for growth and yield modelling and decision support systems to help practitioners make informed decisions (Eid and Hobbelstad 2000, Salminen and Hynynen 2001, Fahlvik et al. 2014).

Traditionally, stand delineation has been a manual and partly subjective process based on spectral and structural cues in aerial images (Avery 1966). In boreal forests, for instance, canopy texture and spectral patterns can reveal dominant species and site conditions (Axelson and Nilsson 1993). Stereoscopic interpretation of overlapping image pairs has commonly been employed to enhance depth perception and improve assessment of canopy structure. However, manual delineation is associated with several well-documented limitations. Shadows and overhanging branches can obscure the ground, leading to misplaced boundaries and biased area estimates (Næsset 1998, Næsset 1999a, Næsset 1999b), and results may vary considerably among interpreters (Nantel 1993, Næsset 1998). These challenges have motivated substantial research efforts towards automated stand delineation methods to improve consistency and scalability.

A wide range of automated approaches has been proposed, with region-growing algorithms dominating much of the literature. Promising results have been reported using high-resolution satellite imagery (Wulder et al. 2008), canopy height models (CHM) calculated from airborne laser scanning (ALS) data (Sullivan et al. 2009), and combinations of spectral and height information (Dechesne et al. 2017). However, as noted by Pukkala (2021), requirements related to stand size, shape, and internal homogeneity differ across management contexts, complicating direct comparisons between methods. Despite numerous attempts at automated stand delineation, large-scale operational implementation has yet to be achieved in Norway (Næsset 2014). Similarly, White et al. (2025) identified automated stand delineation as a key research gap in their review of the status of enhanced forest inventories in Canada.

In a recent study, Sandum et al. (2025) applied a deep-learning approach based on a U-Net architecture (Ronneberger et al. 2015) within a supervised learning framework. Using multispectral aerial images and an ALS-derived CHM as input, combined with a stand map created by an expert interpreter as reference data, they demonstrated that the model could produce delineations comparable to those of a human interpreter. However, model performance declined in structurally complex and heterogenous forest conditions.

One important limitation identified by Sandum et al. (2025) was temporal misalignment between data sources. The aerial imagery and ALS data were collected one year apart, leading to inconsistencies and exclusion of affected areas, thereby reducing the available training data. Such misalignments impose a serious challenge for large-scale applications, where acquisition intervals between data sources may be even longer. Digital aerial photogrammetry (DAP) point clouds offer a potential solution, as point clouds generated directly form aerial imagery provide detailed information about the canopy surface (Baltsavias et al. 2008) and can be used to construct CHMs with height estimates comparable to those from ALS (Mielcarek et al. 2020). By deriving both spectral and structural information from the same image data, DAP ensures temporal alignment between inputs. However, CHMs constructed from DAP point clouds may exhibit smoother canopy surfaces and reduced representation of canopy gaps (Puliti et al. 2015, White et al. 2018). This raises a key research question: can DAP-derived CHMs effectively substitute ALS-derived CHMs when training deep learning models for forest stand delineation? Addressing this question is particularly important given that accurate deep learning models generally require large, high-quality training datasets (Shahinfar et al. 2020, Thian et al. 2022, Chu et al. 2025).

In addition to temporal consistency, guided data set development must also consider inclusion of complementary information that may enhance model performance. Sandum et al. (2025) suggested incorporating a digital terrain model (DTM) could improve delineation accuracy, as terrain information is commonly used by manual interpreters in Norway (Norwegian Agricultural University 1970). Terrain information is especially important in regions with steep topography, where operational constraints strongly influence stand boundaries. To ensure consistency with the reference data, models should therefore have access to comparable terrain information. While the potential value of DTMs for stand delineation has previously been discussed (Leppänen et al. 2008), such information has, to our knowledge, not yet been incorporated into automated stand delineation workflows.

The objective of this study was to evaluate whether CHMs constructed from DAP can serve as a viable alternative to CHMs constructed from ALS data for deep learning-based forest stand delineation, thereby addressing the temporal inconsistencies between data sources. Building on the work by Sandum et al. (2025), we further examined whether the inclusion of a DTM improves the model's ability to replicate manually delineated forest stands.

## 2. Materials and methods

### 2.1 Study area
The study was conducted in the district of Follo in Akershus county, bordering Oslo to the north (Figure 1). The district spans 753 km² and includes the six municipalities: Enebakk, Frogn, Nesodden, Nordre Follo, Vestby, and Ås. Elevation ranges from sea level to 375 m, with a landscape characterized by mixed forests and developed areas. The dominant tree species are Norway spruce (*Picea abies* (L.) Karst.) and Scots pine *(Pinus sylvestris* L.*)*, interspersed with broadleaved species particularly in the coastal zones. The area was selected based on the availability of temporally aligned datasets from 2021, including aerial images, ALS and DAP point clouds, and a forest management plan.

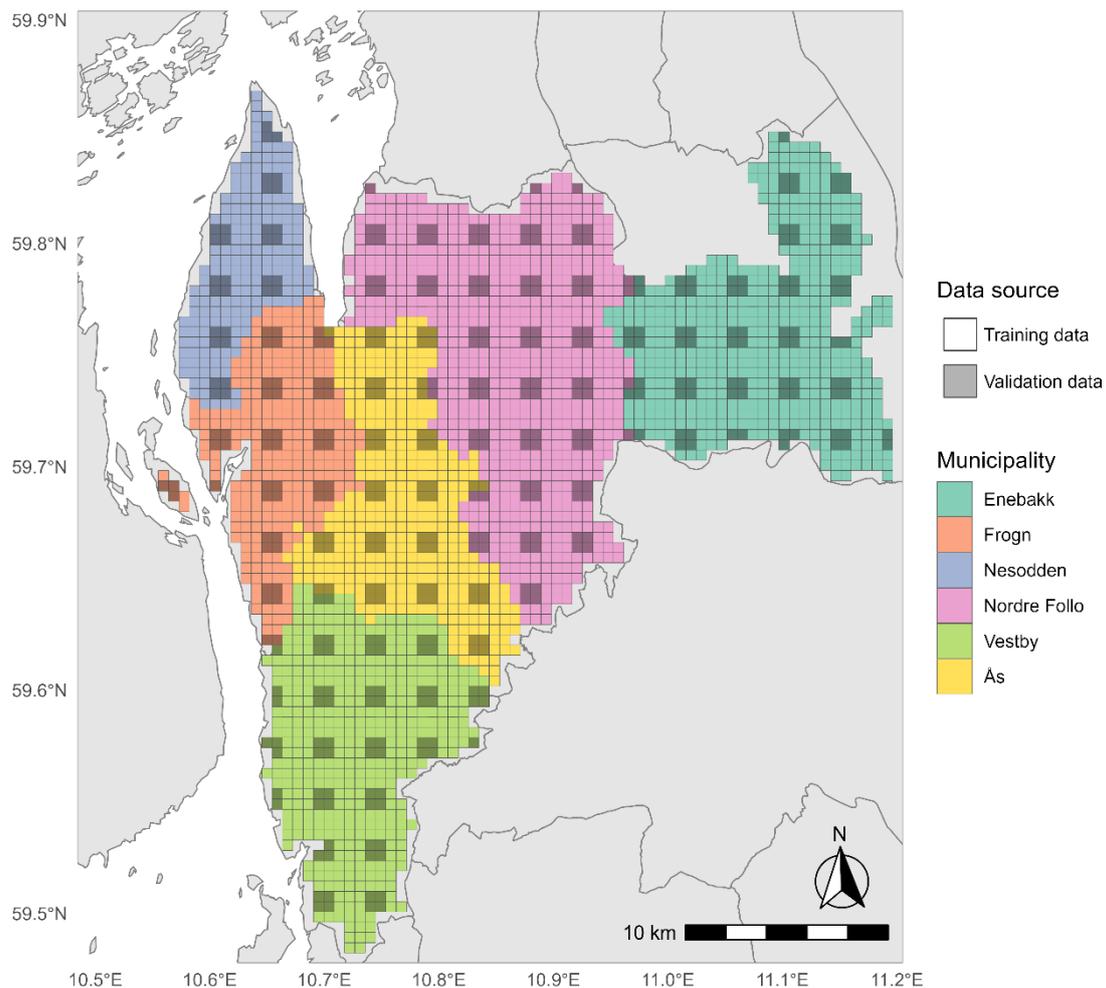

**Figure 1.** *Study area comprising the six municipalities, showing training images in lighter colors and the validation data in darker shades.*

### 2.2 Datasets
The modelling process relied on three primary datasets: ALS data, a DAP point cloud, and a forest management plan containing a stand map. All datasets were acquired in 2021 to ensure temporal consistency. These datasets are described in detail below.

### 2.2.1 Reference data

The stand map, created in 2021 by two expert interpreters with 30 and 40 years of experience, respectively, was produced as part of a forest inventory for management planning. The map was created following a typical Norwegian workflow, consisting of manual interpretation of a combination of stereoscopic aerial imagery, ALS data and an existing stand map (Næsset 2013). The area was first classified as either forested or non-forested. Forested land was then further delineated into individual forest stands and each stand was assigned a development stage.

This classification system divides the forest into five classes: Class I – bare forest land, Class II – recently regenerated forest, with trees up to approximately 8–9 m tall, Class III & IV – young and old thinning-stage forests, respectively, and Class V – mature forests (Norwegian Agricultural University 1970). These classes correspond to potential management actions and are set based on forest age, site index, and tree species. In Class I, regeneration measures should be considered. In Class II, the need for pre-commercial thinning is evaluated. In Classes III and IV, thinning treatments are considered, and Class V indicates forest ready for harvest.

These classes were used as the basis for creating a reference mask for model training. The creation of the mask is described in section 2.3.1.

### 2.2.2 ALS-point cloud

As part of the 2021 forest inventory for forest management planning, ALS data were acquired. Data collection took place over four days: April 23$^{rd}$, April 26$^{th}$, May 28$^{th}$, and May 29$^{th}$, 2021, and was conducted by Terratec AS. A Riegl VQ-1560II dual-channel sensor was flown at 2100 m above ground level, with a pulse repetition rate of 669 kHz, a scan rate of 230 Hz, and a 20% lateral overlap between lines. The resulting point cloud had a pulse density of 5 pulses/m$^2$.

### 2.2.3 DAP-point cloud

Aerial images were captured on April 18$^{th}$, 2021, under leaf-off conditions with a Leica DMC III sensor flown at an altitude of 2600 m above sea level. The images were captured with a lateral overlap of 30% and a longitudinal overlap of 80%. Each image contained four spectral channels (red, green, blue, and near-infrared) with a spatial resolution of 0.1 m and an 8-bit radiometric resolution A DAP-point cloud covering the study area was https://www.hoydedata.no/generated by the Norwegian Mapping Authority using Trimble Inpho Match-T DSM v.11.0.4 software (Trimble Inc 2025).

The point cloud was generated with a point spacing twice the resolution of the original images, resulting in a point density of 25 points/m$^2$, with integrated spectral information. Normalization of the point cloud was performed using ground points derived from the ALS point cloud and the height normalization function applying the k-nearest neighbor inverse distance weighting algorithm in the lidR package v.4.2.1 (Roussel et al. 2020, Roussel and Auty 2025) in R v4.5.2 (R Core Team 2025).

## 2.3 Pre-processing of data

### 2.3.1 Reference masks

To use the stand map for training and validation, it was rasterized based on the stand development stages using the terra package v1.8-60 (Hijmans 2025) in R. During this process Classes I and II were merged into a single class. This was done because a stand is moved from Class I to II once regeneration is deemed satisfactory. However, this distinction is not reliably detectable in remotely sensed data, as seedlings are often too small to be identified. The resulting raster contained five classes: Class NF for non-forested areas, Class I-II for bare forest land and regeneration forest, Class III and IV for young- and old thinning-stage forest, and Class V for mature forest. The raster was one-hot encoded into five binary layers, each representing the presence and absence of a specific class.

### 2.3.2 CHM construction

CHMs were constructed from the ALS and DAP point clouds using the rasterize canopy function from the lidR package in R, applying the point-to-raster (p2r) algorithm with a sub circle radius of 0.15 m. The output rasters were generated at a 1 m resolution, aligning with the aerial images. Negative and missing values were set to zero to ensure valid height estimates, and the height was capped at 50 m, as values exceeding this threshold were considered outliers. Finally, the CHMs were scaled to a range of [0, 1] by dividing all pixel values with 50.

### 2.3.3 Creation of aerial images

True orthophotos https://www.norgeibilder.no/ were reconstructed from the DAP point cloud. Spectral values (RGB and NIR) were extracted using the pixel metric function in the lidR package in R. Images were generated at a 1 × 1 m spatial resolution by averaging the spectral values within each pixel. Missing pixels – mainly in low-contrast areas – were filled using a 3 × 3 kernel mean filter. Visual inspection confirmed the realism of the reconstructed images. Finally, the images were scaled to a range of [0, 1] by dividing all pixel values by 255.

### 2.3.4 DTM construction

A DTM was constructed using the rasterize terrain function in the lidR package in R, applying the triangulated irregular network (TIN) algorithm to interpolate ground elevations. The raster was generated at a 1 m spatial resolution, aligning with the aerial images and CHMs. Negative and missing values were set to 0, to ensure valid height estimates. The DTM was scaled to a range of [0, 1] by dividing all pixel values by 375 m, reflecting the largest observed value.

### 2.3.5 Data type combination

To finalize the model input, the data was stacked into three distinct raster composites designated as RGBI-ALS, RGBI-DAP, and RGBI-DAP-DTM. The RGBI-ALS and RGBI-DAP configurations both formed five-channel inputs by combining the orthophoto with the ALS-derived and DAP-derived CHMs, respectively. For RGBI-DAP-DTM a six-channel input was created by incorporating the DTM into the RGBI-DAP stack. These three data type combinations formed the inputs for model training; hereafter, trained models are referred to by the names of their respective input data.

### 2.4 Data split and cross-validation

The model requires images of a consistent size. To achieve this, the study area was split into square tiles of 512 × 512 m, corresponding to tiles of 512 × 512 pixels (Fig. 1). In total, 2584 tiles were generated.

For model training, the dataset was divided into training and validation subsets. A 15% validation split was selected as a compromise between preserving sufficient training data and ensuring adequate validation coverage. To achieve this, a systematic arrangement of 2 × 2 clusters was superimposed on the study area (Fig. 1). This layout provided spatially distributed validation tiles while minimizing direct adjacency between training and validation sets. Larger contiguous validation clusters were used to help capture small-scale spatial autocorrelation, reflecting the intended application scenario in which the model processes multiple adjacent tiles.

Each tile was assigned to the municipality covering the largest proportion of the tile's area (Table 1). This enabled a spatially explicit cross-validation procedure in which, for each fold, all tiles belonging to one municipality were withheld as the test set. The remaining municipalities provided both training and validation data. Model performance was then evaluated on the held-out municipality. This process was repeated for all six municipalities, resulting in a six-fold cross-validation scheme.

**Table 1.** *Showing distribution of tiles between datasets and municipalities.*

| Municipality | Training n (% of total) | Validation n (% of total) | All images n (% of total) |
|---|---|---|---|
| Enebakk | 449 (84) | 88 (16) | 537 |
| Frogn | 242 (83) | 49 (17) | 291 |
| Nesodden | 166 (84) | 31 (16) | 197 |
| Nordre Follo | 606 (85) | 111 (15) | 717 |
| Vestby | 381 (84) | 74 (16) | 455 |
| Ås | 324 (84) | 63 (16) | 387 |
| **Total** | **2168 (84)** | **416 (16)** | **2584** |

## 2.5 Model training and hyperparameter tuning

The models were trained using the Adam optimizer (Kingma and Ba 2014). As the training objective, a loss function based on the Tversky index (Eq. 1), known as the Focal Tversky loss (eq. 2) was used. The Tversky index combines true positives (TP), false positives (FP) and false negatives (FN), with α and β controlling the relative weighting of FP and FN, respectively. In addition, the focal Tverksy loss incorporates a focal parameter (γ) to encourage the model to focus more on hard-to-classify examples (Abraham and Khan 2019). The focal Tversky loss was averaged across the five classes (c = 5).

$$\text{Tversky index} = \frac{TP}{TP + \alpha \times FP + \beta \times FN} \quad (1)$$

$$\text{Focal Tversky loss} = \frac{\sum_{i=1}^{c} (1 - \text{Tversky Index}_i)^{1/\gamma}}{c} \quad (2)$$

Deep learning model performance is sensitive to hyperparameter configurations, and default settings cannot be assumed to be optimal. To determine optimal values, hyperparameters were optimized using the Optuna framework (Akiba et al. 2019) with the integrated Tree-structured Parzen Estimator sampler. The search space for each hyperparameter is shown in Table 2 and comprised parameters governing both the model architecture and the focal Tversky loss. For each trial, a combination of the hyperparameters was sampled from the search space and evaluated using the six-fold cross-validation strategy. Model performance was measured by the macro averaged Matthew's correlation coefficient (mMCC; Eq. 3), calculated as the arithmetic mean across the five classes. The mMCC metric was chosen as it is known to be an honest and reliable metric (Chicco and Jurman 2020, Chicco et al. 2021, Chicco and Jurman 2023). A total of 30 trials were conducted in the Optuna framework for each data type combination.

$$\text{mMCC} = \frac{1}{c}\sum_{i=1}^{c} \frac{(\text{TP}_i \times \text{TN}_i) - (\text{FP}_i \times \text{FN}_i)}{\sqrt{(\text{TP}_i + \text{FP}_i)(\text{TP}_i + \text{FN}_i)(\text{TN}_i + \text{FP}_i)(\text{TN}_i + \text{FN}_i)}} \qquad (3)$$

Early stopping was applied within each fold to reduce computation time and avoid overfitting. Training was halted if the validation loss did not improve for 10 consecutive epochs, and the model weights from the best-performing epoch were restored. The mMCC metric on the validation set was used as the optimization objective. The models were trained for up to 50 epochs on a single NVIDIA RTX8000 GPU, requiring approximately 30 days of training time for each of the three data-type combinations. A maximum number of 50 epochs was chosen because preliminary tests showed that the models converged within this range.

**Table 2.** *Hyperparameter search space used in the Optuna framework.*

| Hyperparameter | Type | Interval |
| --- | --- | --- |
| **Training** | | |
| Learning rate | Float | [0.1e-5, 0.1e-2] * |
| Batch size | Int | [4, 8, 16] |
| Number of filters | Int | [8, 64] |
| Kernel size | Int | [3, 5, 7] |
| Dropout | Float | [0.0, 0.5] |
| **Loss** | | |
| Alpha | Float | [0.3, 0.7] |
| Beta | Float | 1 - alpha |
| Gamma | Float | [1, 2] |

* Log-scaled to ensure equal sampling density across orders of magnitude.

## 2.6 Model evaluation

In addition to the mMCC metric, model performance was assessed using a set of quantitative classification metrics derived from a confusion matrix. Overall accuracy (OA; eq. 4) was calculated as a general indicator of performance across all classes. To account for class imbalance and ensure equal weighting of each class, the remaining metrics were calculated as macro-averages and denoted by the prefix *m*. These metrices included intersection over union (mIOU; eq. 5), F1 (mF1; eq. 6), user's accuracy (mUA; eq. 7) and producers' accuracy (mPA; eq. 8). In addition to measuring model performance relative to the reference data, these metrics were also used to quantify inter-model agreement by comparing models pairwise, consecutively treating each model's prediction as the reference.

$$\text{OA} = \frac{\text{TP} + \text{TN}}{\text{TP} + \text{TN} + \text{FP} + \text{FN}} \qquad (4)$$

$$\text{mIOU} = \frac{1}{c}\sum_{i=1}^{c} \frac{\text{TP}_i}{\text{TP}_i + \text{FP}_i + \text{FN}_i} \qquad (5)$$

$$\text{mF1} = \frac{1}{c}\sum_{i=1}^{c} \frac{2 \times \text{TP}_i}{2 \times \text{TP}_i + \text{FP}_i + \text{FN}_i} \qquad (6)$$

$$\text{mUA} = \frac{1}{c}\sum_{i=1}^{c} \frac{\text{TP}_i}{\text{TP}_i + \text{FP}_i} \qquad (7)$$

$$\text{mPA} = \frac{1}{c}\sum_{i=1}^{c}\frac{\text{TP}_i}{\text{TP}_i + \text{FN}_i} \tag{8}$$

While quantitative metrics provide an objective performance measure, the quality of segmentation maps cannot be fully captured by these values alone. In particular, stand delineation involves inherent subjectivity in boundary placement. Moreover, confusion-matrix-based metrics treat all misclassifications equally, although in practice, some errors are more severe than others. For example, predicting Class I-II for an area that truly belongs to Class V is more consequential than misclassifying Class III as Class IV. To address these limitations, the predicted segmentation maps were also visually inspected. Three representative examples, selected to illustrate the models' strengths and weaknesses, are presented in the results section. Example 1 was selected because it includes a diverse mix of forest conditions and land-cover types and was deemed representative of model performance. Example 2 was selected to illustrate delineation results across different boundary types. Example 3 was selected to illustrate how dissolving the pixel-wise semantic segmentation masks can, in some cases, produce stand boundaries with high geometric complexity.

## 3. Results

**3. 1 Hyperparameter tuning**
Following the hyperparameter tuning, the single best model for each data combination – determined by the mMCC averaged across folds – was selected for further evaluation. The hyperparameters of these models are presented in Table 3. Across the three datasets, the selected models varied in the number of filters, kernel size, batch size, and dropout level. Notably, the training of the RGBI-DAP-DTM models was frequently halted by the integrated early-stopping procedure due to the validation loss plateauing, whereas the other data-type combinations mostly trained for the full 50 epochs.

**Table 3.** *Results of the Optuna hyperparameter search, detailing the specific configurations that yielded the best performance for each of the three data-type combinations.*

| Data combination | Inference time (mm:ss) | Model parameters | | Training parameters | | | Loss function parameters | |
|---|---|---|---|---|---|---|---|---|
| | | # Filters | Kernel size | Batch size | Learning rate | Dropout | alpha | gamma |
| RGBI-ALS | 18:36 | 64 | 7 | 8 | 1.36E-05 | 0.08 | 0.49 | 1.39 |
| RGBI-DAP | 7:44 | 54 | 3 | 16 | 5.74E-05 | 0.00 | 0.62 | 1.46 |
| RGBI-DAP-DTM | 15:09 | 49 | 3 | 4 | 5.18E-05 | 0.05 | 0.43 | 1.49 |

**3.2 Evaluation metrics and inter-model variability**
Performance across all evaluation metrics for the three input data combinations is summarized in Table 4. Variability across municipalities was low for all three models, indicating stable results on the independent test set. Averaging the performance metrics across all folds showed largely similar values between the different input data combinations, with the models trained on the RGBI-DAP data yielding slightly larger values. Overall accuracy, the only metric not macro-averaged, was substantially larger than the other metrics, showing the importance of macro-averaging metrics. Assessment of inter-model variability by evaluating model predictions against one another yielded larger values than those obtained from comparison with the reference data. This indicated that the models achieved greater consensus with each other than with the reference data, despite independent training on different input data.

Table 4. *Performance metrics from municipality-level cross-validation of U-Net segmentation of forest stands using different input datasets. The top section of each dataset block reports model performance against the reference data across the six municipalities, including the arithmetic mean. The lower section represents a measure of inter-model variability where the model of the current section is compared to the other models used as a stand in for the ground truth. Standard deviations are given in parenthesis.*

| Dataset | Municipality | OA | mMCC | mIOU | mF1 | mUA | mPA |
|---|---|---|---|---|---|---|---|
| RGBI ALS | Enebakk | 0.88 | 0.57 | 0.49 | 0.63 | 0.66 | 0.63 |
| | Frogn | 0.91 | 0.62 | 0.53 | 0.68 | 0.67 | 0.68 |
| | Nesodden | 0.90 | 0.60 | 0.52 | 0.66 | 0.68 | 0.65 |
| | Nordre Follo | 0.90 | 0.60 | 0.51 | 0.66 | 0.66 | 0.68 |
| | Vestby | 0.91 | 0.60 | 0.51 | 0.65 | 0.65 | 0.67 |
| | Ås | 0.93 | 0.64 | 0.54 | 0.68 | 0.68 | 0.69 |
| | **Mean** | **0.91 (0.02)** | **0.60 (0.02)** | **0.52 (0.02)** | **0.66 (0.02)** | **0.67 (0.01)** | **0.67 (0.02)** |
| *Agreement* | **RGBI-DAP** | **0.96 (0.01)** | **0.84 (0.02)** | **0.76 (0.03)** | **0.86 (0.02)** | **0.86 (0.02)** | **0.86 (0.02)** |
| | **RGBI-DAP-DTM** | **0.96 (0.01)** | **0.84 (0.02)** | **0.77 (0.02)** | **0.86 (0.01)** | **0.87 (0.01)** | **0.86 (0.01)** |
| RGBI DAP | Enebakk | 0.89 | 0.59 | 0.51 | 0.66 | 0.67 | 0.66 |
| | Frogn | 0.91 | 0.61 | 0.53 | 0.67 | 0.67 | 0.68 |
| | Nesodden | 0.90 | 0.61 | 0.52 | 0.67 | 0.69 | 0.65 |
| | Nordre Follo | 0.90 | 0.61 | 0.52 | 0.67 | 0.67 | 0.69 |
| | Vestby | 0.91 | 0.61 | 0.52 | 0.66 | 0.65 | 0.68 |
| | Ås | 0.93 | 0.66 | 0.56 | 0.71 | 0.70 | 0.71 |
| | **Mean** | **0.91 (0.02)** | **0.62 (0.02)** | **0.57 (0.02)** | **0.67 (0.02)** | **0.68 (0.01)** | **0.68 (0.02)** |
| *Agreement* | **RGBI-ALS** | **0.96 (0.01)** | **0.84 (0.02)** | **0.76 (0.03)** | **0.86 (0.02)** | **0.86 (0.02)** | **0.86 (0.02)** |
| | **RGBI-DAP-DTM** | **0.96 (0.01)** | **0.84 (0.03)** | **0.76 (0.03)** | **0.86 (0.02)** | **0.87 (0.02)** | **0.86 (0.02)** |
| RGBI DAP DTP | Enebakk | 0.88 | 0.59 | 0.51 | 0.66 | 0.67 | 0.66 |
| | Frogn | 0.91 | 0.61 | 0.52 | 0.67 | 0.66 | 0.68 |
| | Nesodden | 0.90 | 0.58 | 0.50 | 0.65 | 0.65 | 0.67 |
| | Nordre Follo | 0.90 | 0.61 | 0.52 | 0.67 | 0.66 | 0.69 |
| | Vestby | 0.90 | 0.59 | 0.50 | 0.64 | 0.64 | 0.68 |
| | Ås | 0.93 | 0.63 | 0.54 | 0.68 | 0.67 | 0.69 |
| | **Mean** | **0.90 (0.01)** | **0.60 (0.02)** | **0.52 (0.01)** | **0.66 (0.01)** | **0.66 (0.01)** | **0.68 (0.01)** |
| *Agreement* | **RGBI-ALS** | **0.96 (0.01)** | **0.84 (0.02)** | **0.77 (0.02)** | **0.86 (0.01)** | **0.86 (0.01)** | **0.87 (0.01)** |
| | **RGBI-DAP** | **0.96 (0.01)** | **0.84 (0.03)** | **0.76 (0.03)** | **0.86 (0.02)** | **0.86 (0.02)** | **0.87 (0.02)** |

## 3.3 Illustrating examples

Example 1 (Fig. 2) illustrates a scene that includes both forested and non-forest areas, and was selected because it was deemed representative of model performance. The NF class is represented by developed areas in the upper part of the image, as well as an agricultural field in the lower left. The forested areas exhibit substantial structural and compositional variability. Class I-II includes clearcuts and regenerating stands at different stages, ranging from bare forest land to patches showing early signs of regrowth. A deciduous stand dominated by birch is visible near the center of the image, adjacent to the agricultural field. The mature forest classes (Class III-V) are represented by both pure spruce stands (e.g. bottom center of the image) and mixed stands with approximately equal proportions of spruce and birch (top center, near the developed area).

Inspection of the CHMs revealed noticeable differences. The CHM calculated from DAP appeared smoother than the one calculated from ALS. In addition, the DAP CHM reduced performance in the deciduous forest stand, where the canopy height estimates appear less reliable.

The model predictions of all three independently trained models appear broadly similar and aligned well with the reference data (Fig. 2e-h); however, several noticeable differences are present. All of the models failed to detect the forest road in the center of the image. There was also a clear disagreement in the placement of boundaries for the uppermost Class I-II stand, where all models expanded the area relative to the interpreter's delineation. This expansion was more pronounced for the two models trained using the CHM calculated from DAP than for the model trained using the ALS-derived CHM. Furthermore, all three models also classified the bottom-center as pure Class IV, whereas the reference map separates the lower portion as Class V. In addition, the model predictions contained several small patches that were not present in the reference data. These patches were below the minimum mapping unit of 0.2 ha commonly used in Norwegian forest inventories. The presence of such patches, showed a tendency of the models to over-segment the landscape, indicating limited generalization into management-relevant units.

The tendencies discussed above, namely, that the predictions are largely correct, while occasionally missing forest roads, exhibiting over-segmentation that results in some polygons being too small, and the models exhibiting a larger degree of agreement than with the reference are reoccurring themes observed across the entire dataset.

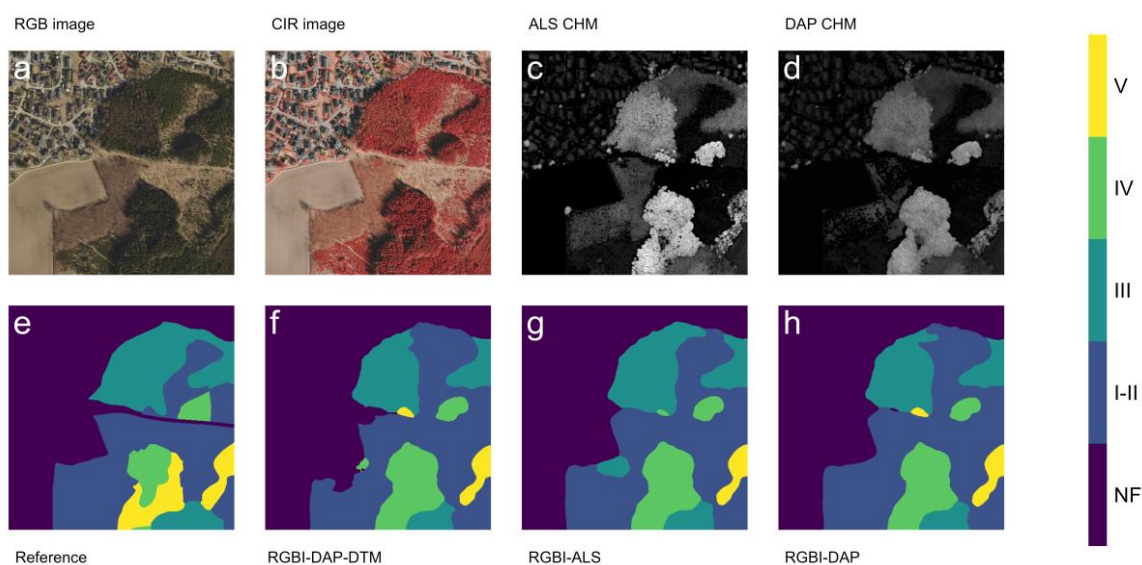

**Figure 2.** *Example 1 showing a single 512 × 512 m tile from Ås municipality, illustrating the input data, reference mask, and the predicted masks. The top row displays the input remote sensing data **(a)** RGB image, **(b)** a color-infrared (CIR) image, **(c)** the CHM calculated from ALS, and **(d)** the CHM calculated from DAP. The bottom row shows **(e)** the reference mask used for validation alongside the model predictions from the three models: **(f)** RGBI-DAP-DTM, **(g)** RGBI-ALS, and **(h)** RGBI-DAP.*

Example 2 (Fig. 3) illustrates delineation results across different boundary types (e.g. clear-cut and Class V or Class III and IV), and the difficulty of evaluating the stand delineation results. From the figure it is evident that the models tended to oversegment the area, producing some small stands that are below the minimum area to be considered independent management units and therefore do not conform to the stand definition. Studying the clear-cuts, the developed area and the agricultural field along top of the image, the models performed well when locating the clear, crisp transitions between these areas and the forest below. A notable exception from this can be seen by studying the regeneration in the lower part of the image. Here, the agreement between the models appears larger than agreement between each model and the reference. All models stretched the regeneration boundary further upwards than the reference. This is in-line with the resulting validation metrics for inter-model variability discussed in section 3.2.

In the older, taller forest areas, where boundaries are less distinct, the four delineations appear similar at first glance. However, closer inspection reveals that even small deviations in boundary placement led to merging and splitting of stands, resulting in substantially different stand configurations overall.

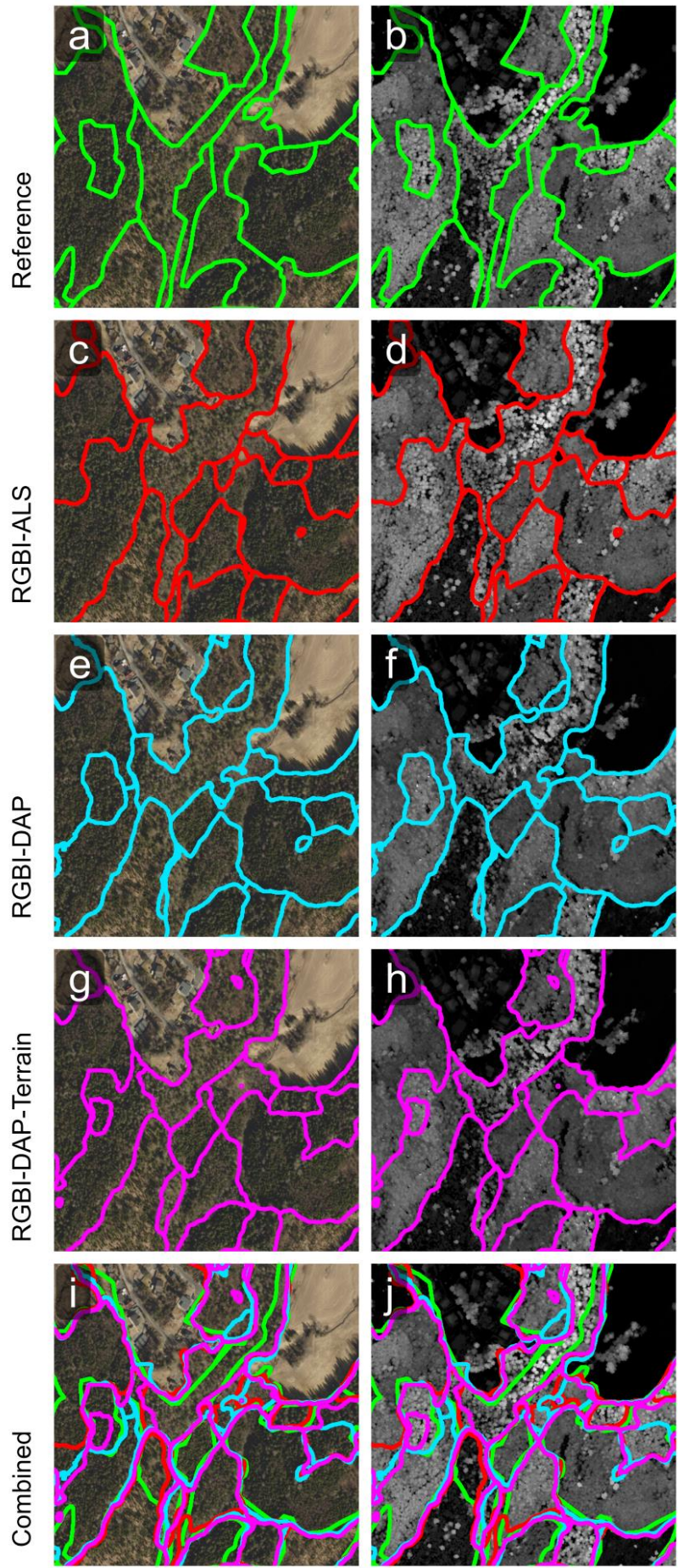

**Figure 3.** *Example 2 showing stand delineation results for a single 512 × 512 m tile in Enebakk municipality. The reference boundaries and model-produced delineations are presented in a two-column format: the left column displays boundaries overlaid on an RGB image, while the right column shows the same results overlaid on a CHM. The rows consist of: **(a, b)** the reference boundary used for evaluation; and the delineations produced by the **(c, d)** RGBI-ALS, **(e, f)** RGBI-DAP, **(g, h)** RGBI-DAP-DTM models. The final row **(i, j)** shows all boundary sets combined to facilitate easy visual comparison.*

Deriving forest stands by dissolving masks from the pixel-wise semantic segmentation masks did in some cases result in stand boundaries with high geometric complexity. This effect was most pronounced in areas lacking clear, well-defined transitions, where gradual changes in forest structure make boundary placement ambiguous. However, complex boundaries were also in some cases observed in areas with visually distinct transitions. Example 3 (Fig. 4) illustrates this by comparing the RGBI-DAP and RGBI-DAP-DTM models. Both models generated fragments too small to constitute separate operational units. Importantly, both models were trained on the same imagery and CHM and applied to terrain with relatively straightforward operating conditions, the RGBI-DAP model produced markedly smoother boundaries than the irregular, complex lines of the RGBI-DAP-DTM model. This could suggest that the observed differences stem from variations in the model training.

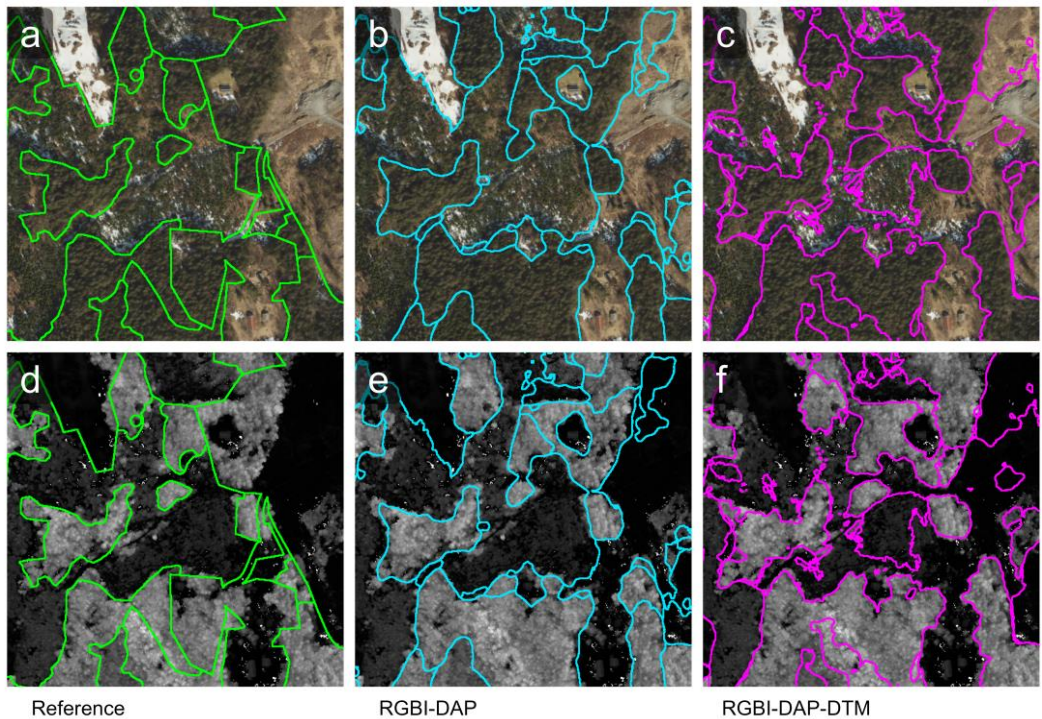

Figure 4. Example 3 illustrating reference and predicted stand boundaries for a single 512 × 512 m tile in Enebakk municipality. The boundaries are overlaid on RGB imagery (top row) and the CHM calculated from DAP data. The panels consist of: **(a, d)** the reference boundaries used for validation; and the predictions produced by the **(b, e)** RGBI-DAP and **(c, f)** RGBI-DAP-DTM models.

# 4. Discussion

This study demonstrated that deep learning-based forest stand delineation can achieve robust and consistent performance across multiple combinations of remotely sensed data. Across all evaluated configurations, the models produced comparable delineations, with mMCC values around 0.60-0.61, indicating a stable segmentation framework that is relatively insensitive to the specific choice of CHM and terrain inputs. Agreement between model predictions generated from different combinations of input data was substantially larger (mMCC = 0.84) than between the model outputs and the reference data, suggesting that the networks learn coherent and internally consistent representations of forest stand structure. This discrepancy also highlights the inherent subjectivity of forest stand delineation, as differences between automated outputs and reference maps partly reflect variation in human interpretation rather than true model errors. Importantly, the achieved performance is comparable to that reported by Sandum et al. (2025), despite notable differences in forest composition, management intensity, and spatial extent, further supporting the interpretation that the models capture a consistent representation of forest stands across contrasting conditions. Remaining discrepancies may therefore arise from ambiguities in the reference data, the quantification of which would require multiple independent representations for the same area and was beyond the scope of the present study. Taken together, these findings indicate that the proposed framework generalizes well across varying conditions and can adapt to the characteristics of the available training data.

A key strength of the present study is the use of temporally aligned remotely sensed data and reference stand maps. By minimizing inconsistencies caused by asynchronous acquisitions – such as harvests or silvicultural interventions present in one data source but absent from others – the analysis enabled controlled comparison between alternative input data combinations. Ensuring temporal alignment of the data constrained the number of viable projects, resulting in the use of data captured during leaf-off conditions, which introduces some specific considerations.

Early spring imagery facilitates discrimination between deciduous and coniferous species due to pronounced spectral differences (Persson et al. 2018). This characteristic is particularly relevant for defining stand boundaries aligning with species composition. However, as seen in Example 1, differences between the ALS- and DAP-derived CHMs were evident, particularly in deciduous stands where photogrammetric matching under leaf-off conditions are known to be less reliable (Baltsavias et al. 2008), and ALS is less affected (White et al. 2025). Nevertheless, model performance was remarkably similar regardless of whether ALS- or DAP-derived CHMs were used. This suggests that the U-Net-based segmentation framework can exploit complimentary information from optical imagery and contextual spatial patterns, even when the elevation signal from the CHM is comparatively weak. From an operational perspective, this robustness is encouraging. In settings where ALS data or high-quality DAP point clouds are unavailable, relying primarily on optical imagery may still yield useful delineations, while reducing preprocessing effort, model complexity, and computational requirements. At the same time, previous studies have shown that CHMs can enhance stand homogeneity and boundary definition in certain algorithmic approaches, indicating that the value of height information is task-dependent and may vary with forest structure, species composition, and management objectives (Mustonen et al. 2008).

The inclusion of a DTM did not lead to improved overall performance but rather resulted in a tendency to overfit, triggered the early stopping criterion and halted the training before the full 50 epochs were completed. This suggests that the additional input increased model complexity without providing sufficiently informative features under the given conditions. Although the Adam optimizer used in this study is generally regarded as a robust default, more sophisticated trainings strategies – such as learning rate scheduling, annealing, or warm-up – have been shown to improve convergence in some deep learning applications (Nakamura et al. 2021), but comes at the cost of increased tuning. The absence of performance gains associated with the explicit terrain information may also be explained by characteristics of the study area and forest management practices. The study area, located in southeastern Norway, is characterized by relatively flat terrain, a limited elevation range, and covers only a single vegetation zone (Moen 1998). Moreover, forest stands are usually managed to maintain their relatively homogenous structure, and original boundaries may partly reflect topographic features. Consequently, the influence of terrain may already be implicitly captured in the RGBI imagery and CHMs, reducing the marginal value of including a DTM. Finally, it should be emphasized that the delineation presented in this study was produced based on semantic segmentation of development stages, which inherently merges adjacent stands belonging to the same class. Reframing forest stand delineation as an instance segmentation problem, in which individual stand instances are explicitly predicted separating adjacent stands that belong to the same development stage, is an interesting direction for future research. This represents a more challenging task and may increase the relevance of additional contextual inputs, including terrain information, particularly in topographically complex landscapes.

No post-processing was applied to the model outputs in order to evaluate the intrinsic performance of the proposed framework. As a result, the predictions occasionally contain small, fragmented regions and locally complex boundaries. From an operational standpoint, such artifacts are not necessarily problematic, as they can be efficiently addressed through established post-processing techniques. Modifying the model's receptive field during training, dissolving small polygons, or applying boundary smoothing can all be used to increase spatial coherence. At the same time, some small regions may represent meaningful features, such as patches of retention trees, and could therefore be of value for biodiversity assessment or certification purposes. This highlights the importance of adapting post-processing strategies to the intended applications.

Despite the potential need for human intervention during post-processing, the proposed approach represents a substantial reduction in manual effort compared with fully manual stand delineation and highlights the promise of deep learning methods for this task. However, the current framework still relies on extensive pre-annotation of substantial parts of the project area, which limits immediate operational transferability. Future research efforts should therefore focus on alternative training paradigms that reduce annotation requirements, such as weakly supervised or semi-supervised learning approaches. These may include coarser annotations (Khoreva et al. 2017), sparse point-based labels (Zhao et al. 2024), or other incomplete reference data that are less costly to acquire than extensive polygon delineations. In addition, semi-supervised strategies such as pseudo-labeling – whereby model predictions on unlabeled data are iteratively used as additional training targets – offer a potential for leveraging large volumes of unlabeled remotely sensed data (Lu et al. 2024). Developing and evaluating such approaches will be critical for improving the applicability of and scalability of deep learning-based stand delineation methods and should be considered a high priority direction for future research.

# 5. Conclusion

In conclusion, the models performed well and demonstrated strong potential for deep learning models in automated stand delineation. No substantial differences were observed between the different combinations of input data, indicating that the model can effectively adapt to the provided input. However, the study area was carefully selected to ensure temporal alignment between data sources, and in operational settings where acquisitions are less synchronized, prioritizing data alignment may be important for optimal performance. Although terrain information did not improve results in this study, it could become more important in complex landscapes or if stand delineation is approached as an instance segmentation problem. Going forward, efforts should focus on reducing noise in predictions, such as fragmented boundaries and small polygons. It will also be important to develop training strategies that support operational deployment while maintaining accuracy.

## Acknowledgements

We sincerely thank Viken Skog SA for preparing and giving access to the forest plan, including the stand map, which was used to train and validate the models presented in this paper.

## CRediT Authorship contribution statement

**Håkon Næss Sandum:** Conceptualization, Methodology, Visualization, Validation, Writing – original draft; **Hans Ole Ørka:** Conceptualization, Supervision, Writing – review and editing; **Oliver Tomic:** Conceptualization, Supervision, Writing – review and editing; **Terje Gobakken:** Conceptualization, Funding acquisition, Project administration, Supervision, Writing – review and editing.

## Disclosure statement

The authors declare that they have no known competing financial interests or personal relationships that could have appeared to influence the work reported in this paper.

## Data statement

The remotely sensed data used in this paper – including the aerial images, the DAP point cloud derived from these images, and the airborne laser scanning data – are owned by the Norwegian mapping authority. The stand map is owned by Viken Skog SA. The authors do not have permission to share either the remotely sensed data or the stand map.

## Funding sources

This work was supported by the Center for Research-based Innovation SmartForest: Bringing Industry 4.0 to the Norwegian forest sector (NFR SFI project no. 309671, smartforest.no).